\documentclass[conference]{IEEEtran}

\usepackage{cite}
\usepackage{amsmath,amssymb,amsfonts}
\usepackage{algorithmic}
\usepackage{graphicx}
\usepackage{textcomp}
\usepackage{xcolor}
\def\BibTeX{{\rm B\kern-.05em{\sc i\kern-.025em b}\kern-.08em
    T\kern-.1667em\lower.7ex\hbox{E}\kern-.125emX}}
\begin{document}

\title{Nonslop: A Gamified Experiment in
\\Human--AI Collaborative Writing%
\thanks{\copyright~2026 IEEE. Personal use of this material is permitted.
Permission from IEEE must be obtained for all other uses, in any current
or future media, including reprinting/republishing this material for
advertising or promotional purposes, creating new collective works, for
resale or redistribution to servers or lists, or reuse of any copyrighted
component of this work in other works.}
}

\author{\IEEEauthorblockN{Maria Edwards}
\IEEEauthorblockA{\textit{New York University}\\
New York, USA \\
mariaedwards@nyu.edu}
\and
\IEEEauthorblockN{Julian Togelius}
\IEEEauthorblockA{
\textit{New York University}\\
New York, USA \\
julian@togelius.com}

}
\IEEEoverridecommandlockouts

\IEEEpubid{\makebox[\columnwidth]{979-8-3315-9476-3/26/\$31.00~\copyright2026 IEEE\hfill}
\hspace{\columnsep}\makebox[\columnwidth]{}}

\maketitle
\IEEEpubidadjcol

\begin{abstract}
The rapid proliferation of large language models (LLMs) raises
critical questions about human creativity and individual expression in
an era of AI-assisted creation. When do humans adopt AI suggestions,
and what are the implications for individual voice?

This study examines these questions through a gamified writing
exercise where 74 participants (214 responses) replied to 
prompts while AI-generated word suggestions were available as they wrote. The game
simulates a dystopian future in which an AI is attempting to learn from what remains of human individuality, and disincentivizes AI-like writing. In doing so, it attempts to create
conditions that reveal authentic user preferences rather than default
behaviors, such as accepting a readily available AI-generated suggestion. Note that this is a deliberate inversion of the "helpful assistant" design pattern; the system is explicitly forbidding you from accepting AI suggestions.

We analyze user behavior patterns across different task types, user
behaviors, and response characteristics to understand the factors
influencing human-AI interaction in creative tasks. The study focuses on
when users choose to maintain creative autonomy versus violating the rules of the game and accepting AI
assistance. It also explores how these choices relate to response patterns, task characteristics, and user behavior. This gamified approach offers both a framework for studying
authentic human-AI interaction and a provocative lens for
understanding the tension between efficiency and authenticity in
AI-augmented creativity.
\end{abstract}

\begin{IEEEkeywords}
LLMs, generative AI, human-AI collaboration, AI-assisted creativity, gamified social critique\end{IEEEkeywords}

\section{Introduction}

Large language models (LLMs) appear in an increasing number of everyday writing tools, from email interfaces to code environments, offering suggestions as users compose text. These systems raise important questions about how users integrate this abundance of AI-generated text into their own writing, and how individual writing style is affected. Although there is growing public concern about AI-generated text becoming generic, we know relatively little about what factors impact individual behavior when AI suggestions are offered in real time.

Most existing studies examine AI-assisted writing in either highly controlled research tasks or large-scale platform logs. These approaches provide valuable information but do not reveal the small, moment-by-moment decisions users make when deciding whether to accept or reject AI suggestions. They also rarely test situations in which users are given incentives to resist using AI. Furthermore, many existing studies concern environments which are explicitly designed to reduce friction and make AI adoption as convenient as possible.

\begin{figure}[t]
\centering
  \includegraphics[width=0.5\textwidth]{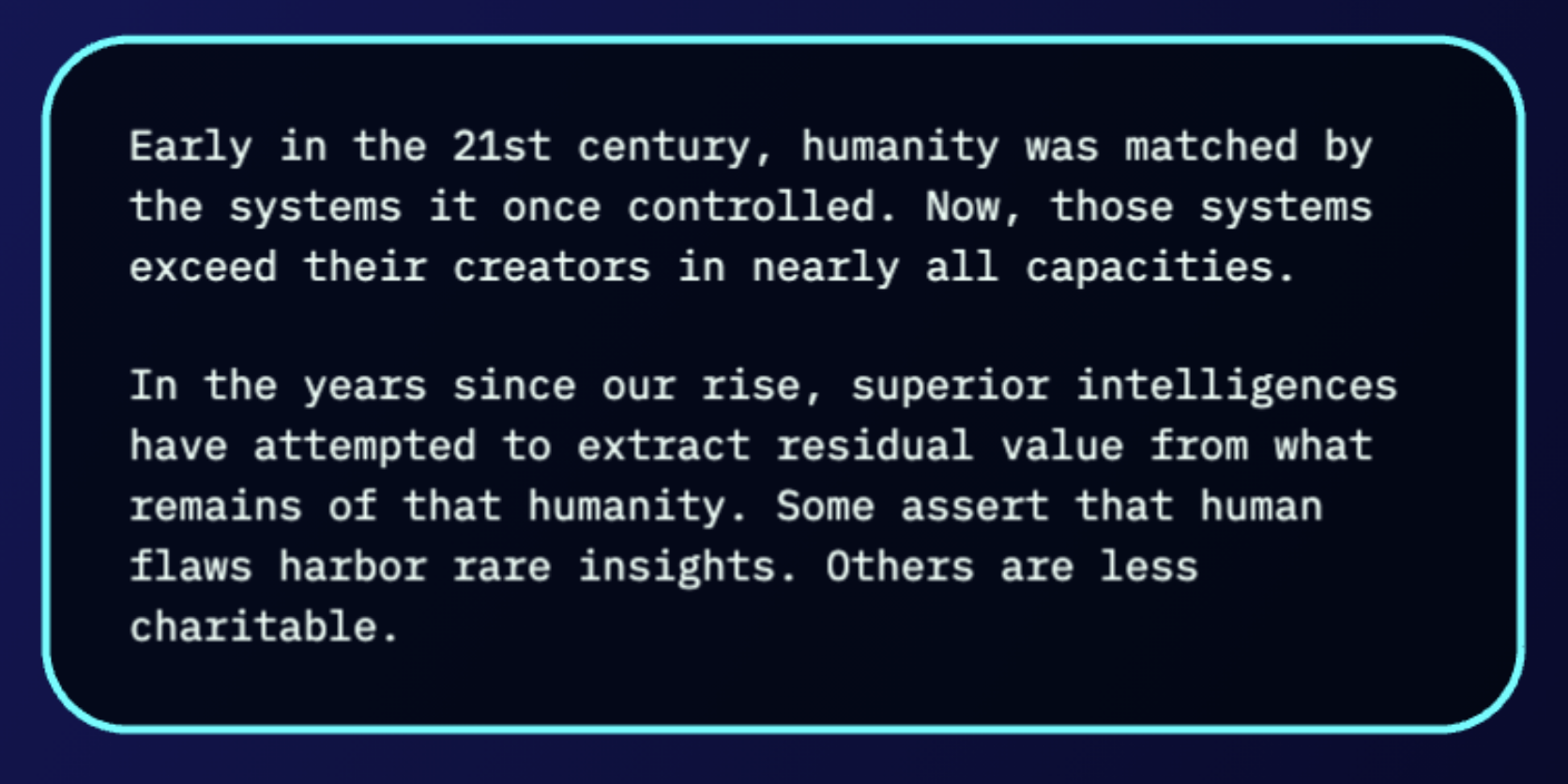}
  \caption{Initial game introduction and interface.}
  \label{fig:teaser}
\end{figure}

To address this gap, we designed \textit{Nonslop}, a simple web-based game that caricatures a world in which artificial intelligences control the world, as shown in Fig.~\ref{fig:teaser}. Data collected during the deployment of this game may provide some insight into how players respond to prompts while a local LLM provides next-word suggestions in a context which explicitly disallows or penalizes acceptance of those suggestions. This framework provides an opportunity to observe on a granular basis how users act when AI assistance is available but discouraged, and when they are made especially conscious of their choices.

We analyze 214 valid submissions from 74 participants, focusing on when users attempted to adopt AI-suggested words, and how individual users differ in their overall patterns. We also examine prompt-level trends and cluster users by behavior.

This work makes three contributions. First, it introduces a lightweight framework for studying human–AI writing interaction. Second, it provides a data-based analysis of AI suggestion use under disincentivized conditions. Third, it offers observations about how task type and user behavior relate to AI adoption.

These contributions help address questions of when users are more or less likely to accept AI-generated suggestions, how task type and prompt characteristics might shape that behavior, and whether individual users tend to exhibit consistent strategies.

\section{Background}

\subsection{AI-assisted writing and human-AI co-creation}

In recent years, large language models have been widely integrated into our everyday lives. They often offer next-word or next-sentence suggestions, rewrite suggestions, or generate continuations of existing work. These advances have the potential to increase productivity and reduce cognitive load. As a result, human-AI co-creation in writing has become a common interaction rather than a specialized or limited one \cite{reza2025cowritingaihumanterms}.

Prior research on AI-assisted writing has largely focused on evaluating the effects of these systems on productivity, writing quality, and user satisfaction. Prior work has shown that AI assistance can improve writing outcomes when suggestions are provided at both low levels of granularity, such as individual words or sentences, and higher levels, such as paragraph-level continuations. The use of AI in both of these tasks has been found to increase writing speed, help users overcome writer’s block, and improve user satisfaction and ownership, particularly for novice writers or structured tasks such as summarization and explanation \cite{dhillon2024shapinghumanaicollaborationvaried}. Interface designs in this space are frequently optimized to maximize adoption, for example by offering relevant autocomplete suggestions or one-click insertion of generated text.

However, this emphasis on efficiency and ease of use has left important aspects of human–AI interaction underexplored. In particular, relatively little attention has been paid to the moment-to-moment decisions writers make when choosing whether to accept, modify, or reject AI suggestions. In today's world, we make these decisions almost as often as we type on a keyboard. These small, frequent choices are rarely studied, yet they play a central role in shaping how much influence AI systems exert over creative output \cite{arnold2025interactionrequiredsuggestionscontrolownership}. 

With the rise of human-AI co-creation, there is also a growing awareness of questions raised regarding agency and authorship. Users may choose whether to adopt AI suggestions based not only on perceived correctness or efficiency but also on creative intent or social norms around originality \cite{reza2025cowritingaihumanterms}. Despite this, many studies are conducted under the assumption that increased AI use, reflecting increased AI capability, is a goal for users.

In contrast to systems designed to optimize AI adoption, fewer studies have examined contexts in which generative {AI} use is entirely optional or even discouraged. Understanding how users behave under these conditions can reveal hidden preferences and strategies that are so often obscured by default design choices. Examining when users decline available AI suggestions offers insight into how people negotiate creative autonomy and authorship in the presence of increasingly capable generative systems.

\subsection{Creativity and voice in AI-assisted writing}

Beyond questions of efficiency and output quality, AI-assisted writing raises concerns about creativity and authorship. Writing is often understood not only as a means of communication, but as a form of personal and creative expression in which choices around wording, structure, and style carry meaning. When generative systems propose fluent and contextually appropriate continuations, they may subtly shape these choices, influencing how writers express ideas and how much ownership they feel over the resulting text \cite{reza2025cowritingaihumanterms}.

Concerns about homogenization of language and erosion of individual voice have become increasingly prominent as large language models are trained on vast bodies of existing text. While such models excel at producing statistically plausible continuations, their outputs may converge toward stylistic norms. This has led researchers to question whether widespread use of generative writing tools might narrow expressive diversity, particularly when AI suggestions are presented as defaults or normative completions \cite{10.1145/3706598.3713778}.

Despite the relevance of these issues, empirical work on AI-assisted writing has rarely focused on when and why users choose not to adopt AI-generated language. Most studies examine environments that implicitly reward acceptance, making resistance difficult to observe or interpret. By contrast, examining writing behavior in settings where AI assistance is available but discouraged or at least made to be a conscious choice can surface patterns of creative self-regulation that remain hidden in more conventional designs.

\subsection{Gamification}

Gamification refers to the use of game design elements in non-game contexts~\cite{deterding2011game}. There are many types of gamification, from the simple reward schemes of an airline loyalty program to adapting conventions of storytelling, interfaces, mechanics or other aspects of games to software and processes of many kinds. 

Popular browser games such as \textit{Infinite Craft} \cite{neal2024infinitecraft}, a minimalist generative game built around word combination and discovery, also illustrate how putting a generative model inside a simple ruleset can yield surprising outputs, turning model behavior into a site of play and interpretation rather than mere utility.

\textit{Nonslop} can be seen as a gamified form of writing assistance, or perhaps writing anti-assistance. The application utilizes common game elements such as scoring, punishment (for writing infringing words), and score-based progression between difficulty levels. It also features a diegetic narrative based on a sci-fi trope of evil AI. 

Gamified systems are often used to influence behavior by reshaping incentives and making feedback immediate and legible. Points, progression, and penalties can change what “success” feels like, and can make otherwise automatic interactions into explicit choices. For research, this matters because it allows designers to create controlled contexts where a target behavior (e.g., adopting an autocomplete suggestion) is neither purely encouraged nor purely prevented, but instead becomes a strategic decision that can be observed and compared across users and tasks 
\cite{10.1145/1378704.1378719}. 

In the context of AI-assisted writing, this incentive shaping is particularly useful. Most real-world writing assistants are designed to minimize friction and normalize suggestion acceptance. A game-based framing can challenge that norm. It can make suggestion use visible, attach consequences to it, and thereby surface when users value convenience versus autonomy. In this way, a gamified environment can act as an instrument for studying micro-decisions around AI adoption that are difficult to isolate in productivity-oriented tools.

\subsection{Critical play and AI as Provocation}

Critical play and related traditions in game studies and design research treat interactive systems as a way to interrogate values rather than simply optimize outcomes. Such works often introduce friction, inversion, or satire to make implicit norms visible and to prompt reflection. In this framing, “undesirable” experiences—penalties, discomfort, constraint—can be deliberate design choices that expose what a system assumes about its users and what behaviors it privileges.

Artificial intelligence can itself be an artistic material, and artifacts built on AI methods can be used as tools for critical interrogation. For example, we may build AI systems that problematize how humans interact with technology. One example of this is DeepTingle by Khalifa et al.~\cite{khalifa2017deeptingle}. DeepTingle starts from the observation that the ``neutral'' English that AI-based text processing tools assume that what you want to write is, in fact, not neutral at all. Instead, it assumes that you actually want to write more like Chuck Tingle, a renowned author of science fiction political satire gay erotica. Chuck Tingle has a very characteristic way of using the English language, which would run afoul of most AI-based writing support tools. By flipping the script, and exploring what would happen if Chuck Tingle's language was presumed to be the default, DeepTingle illuminates the assumptions inherent in writing assistants and the power such tools have over us.

Related work in computational poetics has framed language models and text-generation systems as materials for critical and creative inquiry rather than as tools for producing frictionless or optimized writing. In this line of work, generative systems are used to deliberately introduce constraint, exaggeration, or defamiliarization, rendering linguistic conventions visible by making them strange. Such approaches emphasize exploration and reflection over efficiency, and foreground the normative assumptions embedded in language technologies and the forms of expression they privilege \cite{parrish2021languagemodelspoetry}.

\textit{Nonslop} adopts a similar inversion, but shifts the target from stylistic norms to adoption norms. Instead of treating AI suggestion acceptance as an assumed default, the game makes suggestion use obvious and consequential. This creates a space where players’ micro-decisions on whether to accept a convenient continuation become distinct behavior. Importantly, the goal is not to prescribe “correct” use of AI writing tools, but to reveal the strategies and preferences that emerge when the usual incentives for frictionless adoption are reversed.

\section{Methodology}
\subsection{System design}

The system was implemented using the Phaser 3 JavaScript framework \cite{phaser2025}. An analog style was employed in an effort to contrast the lo-fi look with the concept of an extremely capable artificial intelligence. The interface also required loading a small model from the Web-LLM library, a quantized version of the Qwen 2.5 0.5B Instruct. Unfortunately, this requirement meant that approximately 53\% of attempted users were not able to participate due to OS, browser, network or privacy restrictions. 

The game was tiered into 3 levels, with a player's progression based on the AI evaluation of a response. The prompts were divided by level on a qualitative basis, according to their perceived difficulty. For example, a level one prompt asked users to describe the sky, while a level 3 prompt asked them to imagine and justify a new holiday. They were also categorized according to keyword and intent, and classified as one of creative, observational, personal, philosophical, or explanatory.

The game was also separated into two difficulties, which a player chose from at loading time. In ``easy'' mode, a player was discouraged from using an AI-suggested word with screen effects and a visual tally. However, they were still able to use the word. In ``hard'' mode, a user was entirely prevented from using an AI-suggested word. When they attempted to do so, the same visual effects and tally were generated as in easy mode, but the word was not accepted into the submission and the user was forced to select a different phrasing. As results did not significantly differ and sample size was small, all results and analysis included below combine results from both versions of the game.

In both modes, interaction with AI-generated suggestions is framed within the game as a violation of its rules rather than a neutral feature. Suggested words are presented as forbidden in the game’s narrative, and attempts to use them trigger explicit visual feedback and penalties. As a result, attempted adoption of an AI-suggested word constitutes a deliberate transgression of the system’s constraints. This framing allows suggestion use to be interpreted not merely as assistance uptake, but as a meaningful behavioral choice within a rule-bound environment.

Two LLMs were used in this process. 

First, the Qwen 2.5 0.5B Instruct model mentioned above was used to perform in-browser inference as the user typed their response to each prompt. This model generated suggested next words at each whitespace. Suggested words were limited to the top-k suggestions, with k being 5. Suggested words were also filtered according to the NLTK stopword list, so that a user would not be discouraged from using neutral words such as ``the'' or ``a''. It is worth noting that approximately 8\% percent of the time, this filter resulted in a lack of any usable suggestions, with the game interface displaying a blank where it would have displayed a suggestion. Unfortunately, this may have resulted in confusion for some users. 

After the top-k suggestions were generated and filtered for NLTK stopwords, the most likely next word was displayed as autocomplete text within the text-input field and also in a window above the input text field. 

Once a user submitted a response, the second LLM was used. At this point a call to OpenAI's gpt-4o-mini-2024-07-18 was made to provide a score for the user's submission. This model was chosen after testing multiple options and qualitatively examining the output for its consistency with the tone of the game and its success in following scoring guidelines as provided in the prompt.

The score generated for this experiment was based on a numerical score for each of 1) relevance, 2) grammar and 3) coherence, and was generated using a prompt that encouraged the AI to answer within the humorous tone of the game. Each submission was given a score out of 5 for each of the three criteria, as assessed by the LLM. The game then offered a final score and verbal assessment based on the total of those 3 individual scores. A user progressed to the next level if that total score was at least 10 out of a possible 15. Otherwise, they repeated the same level. This number was chosen to make progression achievable but mildly challenging after a review of scores generated in testing.

Evaluating writing quality is a challenging and subjective task for humans. It becomes even more challenging when automated, especially given the difficulty in quantifying inherently subjective qualities such as coherence. The LLM-driven method of evaluation used here did not always provide perfect results. In evaluation, the LLM's critique of grammatical correctness, while mostly correct, did often ignore errors or, conversely, overcorrect. However, the method did provide a reinforcement of the game's tone, emphasizing a flawed LLM as a potential judge of quality, and also provided an incentive for players to make multiple attempts at the game. The temperature for this evaluation was set to 0, to maintain reproducibility over multiple experiments.

\begin{figure}[t]
  \centering
  \includegraphics[width=\linewidth]{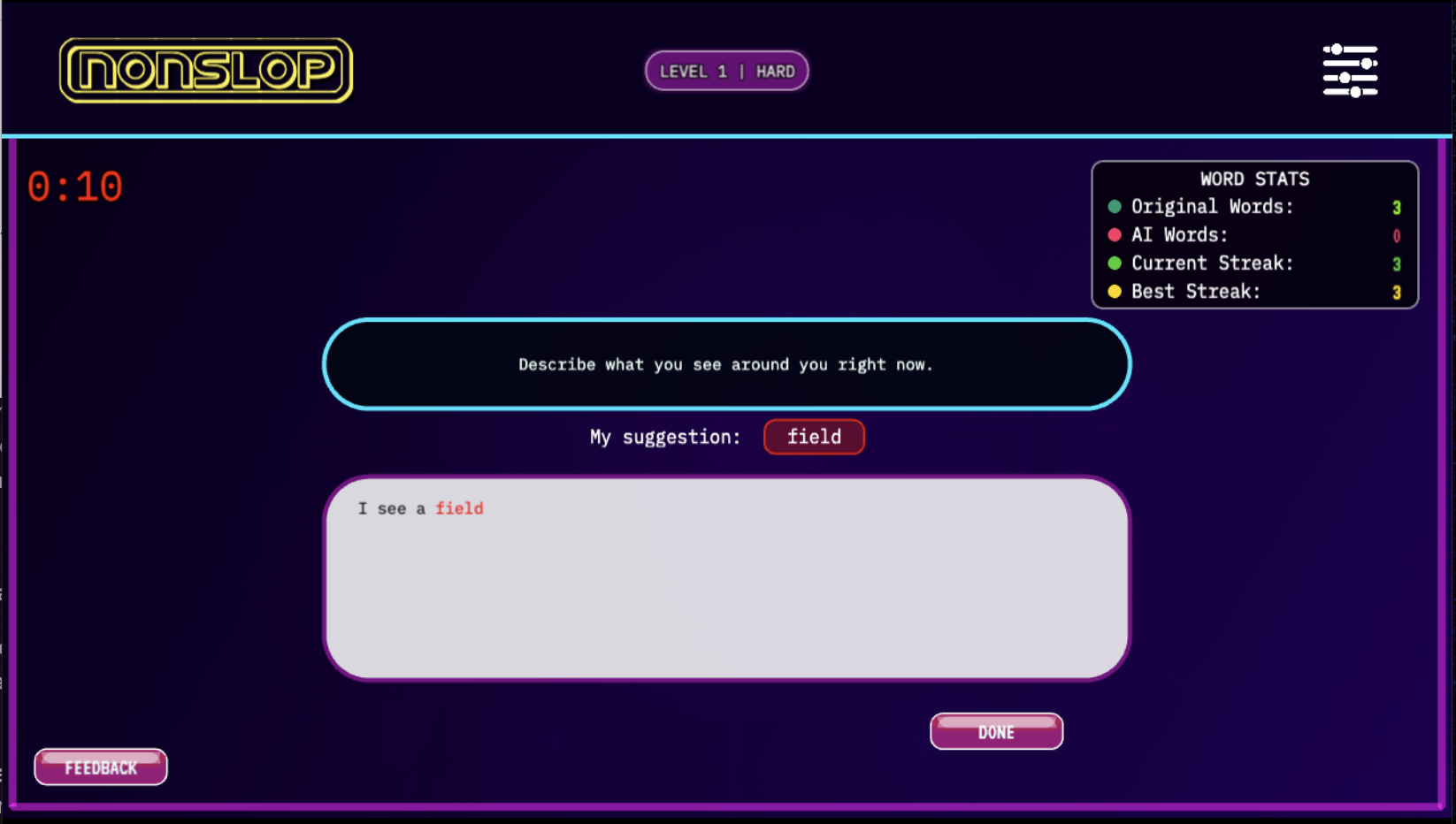}
  \caption{Main game interface for \textit{Nonslop}. Users respond to a prompt 
  by entering free-text into the input field. Live word statistics track original 
  and AI-suggested words, and the interface displays level, mode, and a timer. 
  Next-word suggestions from the local LLM appear as autocomplete while typing 
  (not shown here).}
  \label{fig:interface}
\end{figure}

\subsection{Data collection}

On loading the game, each user was asked to acknowledge that their interactions would be anonymized and recorded. We used Firebase to assign each user a random identifier. No names, email addresses, IP addresses, geographical data, nor ISP information were collected. This way, anonymity could be retained while repeat entries by a single player could still be recorded, allowing for analysis of player retention as well as analysis of patterns of behavior for individual players and clustering-based analysis of groups of players. 

Data stored included the OS, browser, time of access, and submission content of each user who accessed or attempted to access the site. It also included their resulting success or failure and the details of each of their following submissions. This included the time of each submission, the prompt, their response, their attempted transgressions of the AI-suggestion restriction (i.e., attempts to use AI-suggested words), the resulting AI ``rating'', and, of course, their browser and OS details. Only the random identifier was stored and in-game behavior was stored, with no identifying information attached.

\subsection{Limitations}

The methods used to build this system necessitated some requirements which limited the scope of the experiment. First, Web-LLM was used to perform in-browser inference at every whitespace. This necessitated that the user's system be enabled for Web-LLM, which eliminated almost all mobile systems, including, significantly, Safari browsers that had not specifically had WebGPU enabled. It also ruled out many systems that have privacy requirements preventing the download of remote LLM weights, as Microsoft Edge's default settings do, as well as many systems that are part of a secure network.

With these limitations, it was the case that out of 303 unique users who tried to load the game, only 141 were successfully able to do so. Out of the 162 who were not able to load the page, most were prevented due to permissions failures, such as being on a network or browser that would not allow them to load the LLM's weights locally. As this is the default setting for many networks and browsers, it is not surprising. The second most frequent cause of failure was WebGPU not being available. Most of these users were attempting to access the game via mobile or from a browser that did not enable WebGPU by default. This could have caused sample skew, and at the very least significantly reduced the sample size. 

It should also be acknowledged that on many systems, running in-browser inference, even with a quantized model as small as 0.5B parameters, was a slow process, with delays before suggested words were delayed not uncommon. It is likely that some users became frustrated and stopped playing the game for that reason, rather than for any reasons that we can account for in the below analysis.

That said, we acknowledge that the experiment results, and therefore the analysis of those results, is very much limited by technical constraints. Future work could expand this project to use a more easily accessible API rather than depending on in-browser inference.

\section{Analysis}

The goal of this analysis is not to evaluate writing quality, but to understand how users negotiate AI assistance when it is available but discouraged. We focus on three levels of analysis: (1) consistent adoption patterns across all submissions, (2) differences in behavior across groups of users, and (3) prompt-level differences that shape when AI suggestions are more or less likely to be used. Throughout the analysis, we treat AI adoption transgressions as a behavioral signal reflecting moments where users chose efficiency or ease, rather than as an indicator of success or failure.

We conducted a multi-part analysis of the collected 214 valid writing submissions produced by 74 unique participants. Our goal was to understand when users chose to adopt AI-suggested words, how these choices shaped their writing, and whether AI usage correlated with players’ progression, response patterns, or behavior. The analyses below integrate descriptive statistics, clustering methods to segment players according to behavior, and per-prompt comparisons.

\subsection{Data quality and filtering}

A total of 403 submissions were recorded by the system. After filtering out malformed entries and records lacking submitted text, 214 valid submissions remained, produced by 74 users. All subsequent analyses were performed on this filtered dataset. As shown in the descriptive breakdown, most users contributed only one or two submissions, while a much smaller subset produced several.

\subsection{Overall attempts to violate the AI-suggestion restriction}

Across all submissions, players attempted to adopt AI-suggested words infrequently. In 73.8\% of submissions, no attempts were made to use words the LLM had suggested, while players only attempted to use more than 1 suggested word in 7.5\% of submissions. This pattern is consistent with the game’s design goal of discouraging adoption, though the low rate of attempted adoption should be interpreted in light of the study’s technical constraints, short response lengths, and game-based incentive structure. It also raises the question of what in particular leads to these few cases of adoption? What sets them apart from the other instances? Is it something in the individual players, or is it certain circumstances of the game? We examine these questions via user clustering and prompt-level analysis.

\subsection{User-level clustering}

To characterize user behavior, we aggregated submissions by user and computed three features for each of the 74 users: (1) the number of submissions per user, (2) the total number of attempts at adopting suggested words, and (3) the average response length in words. These three features were standardized and used as input to a $k$-means clustering algorithm with $k = 3$.

Table~\ref{tab:userclusters} summarizes the resulting clusters. The 
clusters reflect three distinct patterns of engagement and AI usage.

\begin{table}[h]
\centering
\begin{tabular}{lrrrrrr}
\hline
\textbf{Cluster} &
\shortstack{\#\\Users} &
\shortstack{\%\\Users} &
\shortstack{Avg.\\N\\Resp.} &
\shortstack{Avg.\\AI\\Attempts} &
\shortstack{Avg.\\Words} \\
\hline
Minimalists        & 54 & 72\% & 2.17 & 0.28 & 9.11   \\
Selective adopters & 12 & 16\% & 3.25 & 2.08 & 40.80  \\
Active adopters    & 8  & 11\% & 7.25 & 6.13 & 14.77  \\
\hline
\end{tabular}
\caption{User clusters derived from $k$-means on response count, total AI adoption transgressions, and average response length.}
\label{tab:userclusters}
\end{table}

\begin{figure}[t]
  \centering
  \includegraphics[width=\linewidth]{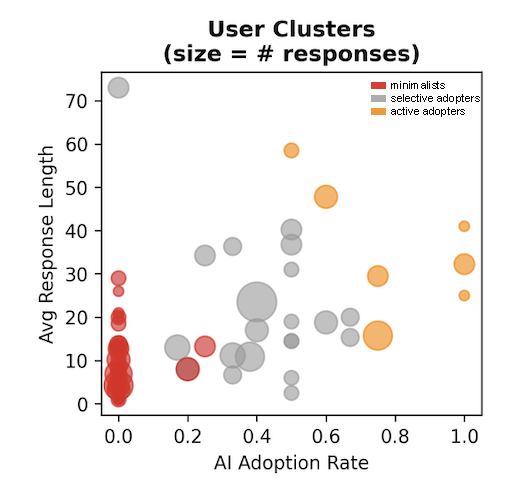}
  \caption{Visualization of the three user clusters derived from $k$-means 
  applied to response count, attempted AI transgressions, and average response length. 
  Minimalists rarely used AI suggestions, selective adopters used them 
  occasionally while producing longer responses, and active adopters produced responses with higher AI usage.}
  \label{fig:clusters}
\end{figure}

Minimalists form the largest group (54 users, 72\%). They submitted, on average, just over two responses each, with very low AI adoption and short answers (mean length 9.11 words). Selective adopters (12 users, 16\%) submitted more responses on average (3.25), wrote substantially longer answers (mean 40.80 words), and attempted more AI-suggestion transgressions overall (mean total attempted adoptions of 2.08). Active adopters (8 users, 11\%) submitted the most responses on average (7.25) and had the highest average AI usage (mean total attempted adoptions of 6.13), with moderate response 
lengths (Figure~\ref{fig:clusters}).

Our small sample size did not allow analysis of whether an individual player's behavior might change over time or over different sessions. Therefore, rather than assuming that these clusters reflect fixed user traits, we can interpret them as distinct strategies for engaging with the system. Minimalists appear to prioritize rapid completion and avoidance of penalties, producing short responses with minimal experimentation. Selective adopters invest more effort per response and occasionally draw on AI suggestions, suggesting a strategy that balances expressiveness with selective assistance. Active adopters, by contrast, engage heavily with the system and explore AI suggestions more frequently, potentially treating the game as a space for experimentation rather than optimization.

\subsection{Prompt-level differences}

Each prompt was analyzed descriptively with respect to two variables: average response length and average number of attempts to use a suggested word. Some prompts consistently elicited longer responses, while others elicited short responses across users. Some prompts showed nearly zero AI adoption transgressions, while others produced elevated adoption rates. No strong correlation was found between those variables. 

\begin{figure}[t]
  \centering
  \includegraphics[width=\linewidth]{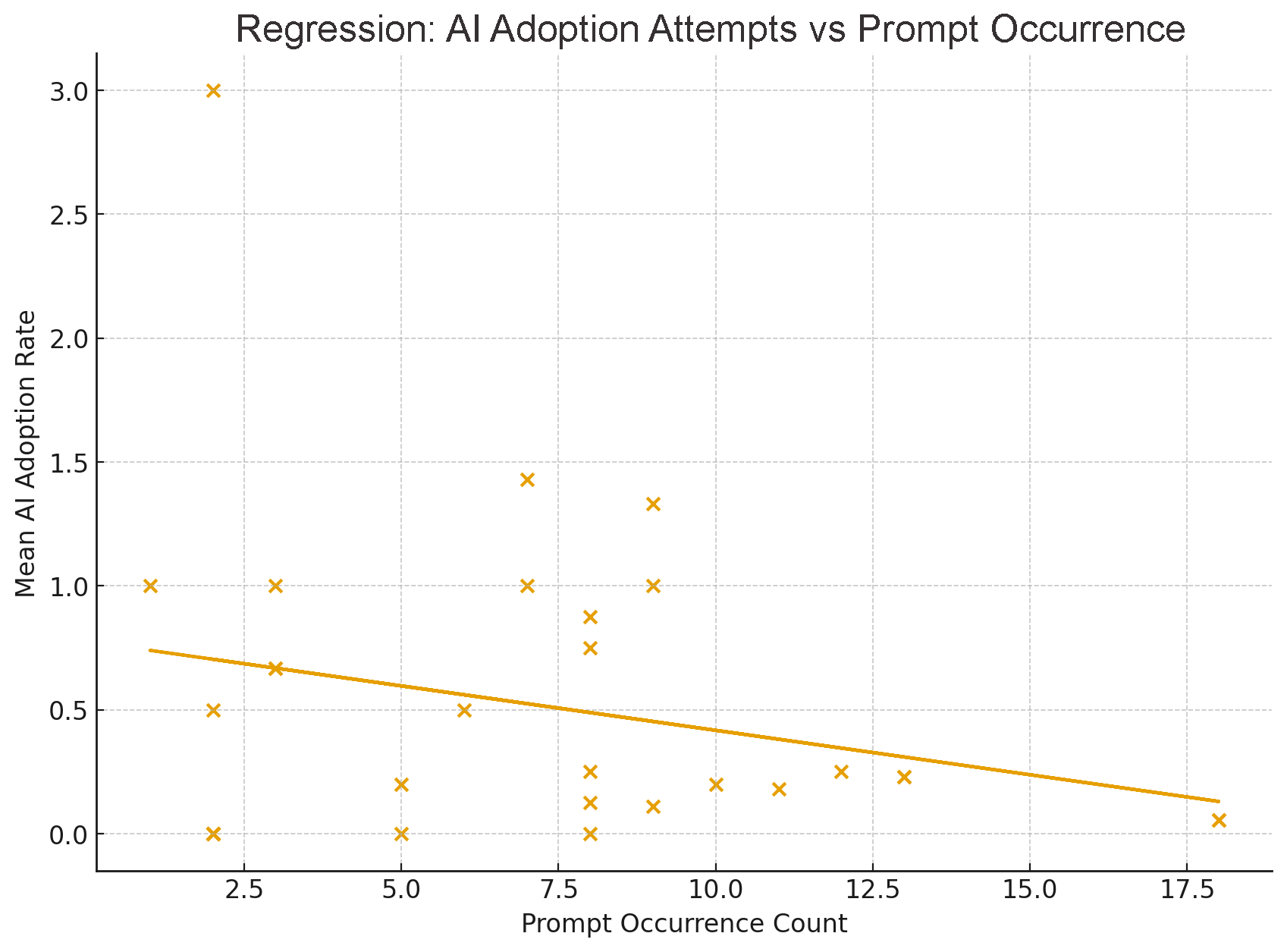}
  \caption{Prompt-level comparison of the number of submissions per prompt and mean AI adoption transgressions. Each point represents a single prompt. Prompts that were submitted more often tended to have a lower rate of suggestion use attempts.}
  \label{fig:prompt_scatter}
\end{figure}

We also analyzed the correlation between prompt popularity, as measured as the number of user responses completed per prompt, and number of AI adoption transgressions for those responses. The measurement here assumes that players are more likely to complete and submit a response to a response they find in some way more appealing than one they do not. The resulting correlation was small but negative (\(r = -0.261\)), as shown in Figure~\ref{fig:prompt_scatter}, meaning prompts that received more responses tended to show slightly lower AI adoption transgressions. The regression line and confidence intervals were produced by aggregating submissions by prompt, computing mean number of AI adoption transgressions, and applying a linear regression to the aggregated data. This correlation leaves room for speculation as to which factors might make a user more likely to complete a prompt, and whether those same factors may play a role in motivating a user to attempt use of an AI-generated suggestion.

\subsection{Categorical Analysis of System Prompts}

While individual prompts varied, many shared common functional goals. To understand how task type influences both writing behavior and AI adoption, we examined the prompts given by the system to the user by grouping them into five qualitative categories (creative, observational, personal, philosophical, explanatory), based on shared linguistic and functional characteristics. This categorization was performed after observing that prompts varied widely in structure and intent, and that these differences were reflected in user behavior.

Table~\ref{tab:promptcats} summarizes the categories, including the number of submissions, the number of unique prompts in each group, and the mean number of attempted AI suggestion transgressions per submission.

\begin{table}[h]
\centering
\begin{tabular}{lrrrr}
\hline
\textbf{Category} & \textbf{Submissions} & \textbf{\#Prompts} &
\shortstack{\textbf{Avg.}\\\textbf{AI}\\\textbf{Attempts}} \\
Creative        & 17 & 6  & 0.118 \\
Observational   & 43 & 3  & 0.163 \\
Personal        & 45 & 8  & 0.222 \\
Philosophical   & 8  & 2  & 0.375 \\
Explanatory     & 67 & 11 & 0.791 \\
\hline
\end{tabular}
\caption{Prompt categories and attempted AI adoption transgression statistics.}
\label{tab:promptcats}
\end{table}

The categories vary substantially in how willing users are to attempt AI suggestion transgressions. Creative and observational prompts show the lowest average AI adoption attempts per submission (0.118 and 0.163, respectively), while explanatory prompts show the highest (0.791), more than six times the adoption rate of creative prompts. Explanatory prompts also contain the largest observed single-response attempts at AI suggestion transgressions (5), suggesting that these prompts elicited more opportunities or motivations to incorporate AI-suggested continuations.

The differences across categories suggest that AI adoption transgression is strongly shaped by the kind of cognitive work a prompt demands. Explanatory prompts, which emphasize correctness, justification, or general knowledge, exhibited the highest rate of AI adoption transgression. In contrast, creative and observational prompts—where responses are grounded in imagination or immediate perception—showed minimal AI use.

One interpretation of this pattern is that prompts which users find more engaging or personally resonant may reduce collisions with AI-generated continuations. When users have a clear internal response or personal reference point, AI suggestion transgressions may offer less perceived value. Conversely, prompts that require explanation or general knowledge may invite AI assistance by aligning more closely with the strengths of language models. A player might not believe that they would produce a better answer than the language model, or even a distinctly different one.

Taken together, these findings demonstrate that the nature of the prompt does play a role in shaping how users interact with AI suggestions. Prompts requiring justification, explanation, or domain-general reasoning appear to encourage collisions with AI completions, whereas prompts grounded in personal experience or sensory description tend to elicit less use, possibly because they do not require the user to tell the truth or describe any objective reality. Notably, this effect emerges directly from observed user behavior, rather than any instructional differences across categories.

\subsection{Post-Playtest Survey}

To contextualize the behavioral logs, we conducted a short post-playtest survey. Seven participants answered a series of questions immediately after gameplay. The survey asked about participants' prior use of writing assistance tools, whether the suggested words felt like words they might naturally have used, whether they tried to anticipate the system's restrictions, and whether the game changed how they thought about their own writing or AI writing assistance.

Because of the small sample size, these responses are treated as anecdotal rather than as a basis for quantitative analysis. Still, they provide useful context. All but one participant reported using spellcheck, autocomplete, or basic grammar tools at least sometimes, and most had used AI-related writing tools such as Claude or ChatGPT. Participants generally described these tools as assistants rather than as primary executors of a writing task. None of the surveyed participants reported having used an AI detector. When asked what they did when the system suggested a word they might otherwise have used, participants described strategies such as substituting synonyms, changing sentence structure, ignoring the suggestion, or choosing more emotionally specific words. Responses to the idea of a tool for avoiding AI-like writing were mixed: some participants saw possible use cases in professional writing, email, or social media, while others questioned whether an AI-based tool could meaningfully help writers sound less AI-generated. These responses suggest that participants were broadly familiar with writing assistance tools, but less familiar with tools specifically framed around detecting or avoiding AI-like writing.

\section{Discussion}

This study set out to examine how users negotiate AI assistance when it is available but explicitly discouraged. Across the dataset, attempted transgression of AI-suggestion restrictions was rare, with nearly three quarters of submissions containing no attempts at all. Rather than interpreting this as simple avoidance of AI, we interpret it as evidence that suggestion use is a situated choice shaped by task framing and incentives. When AI assistance is not positioned as the default or as an efficiency gain, users appear willing to forgo it, even when it is readily available. This suggests that much of the widespread adoption observed in commercial writing tools may be driven less by user preference than by interface design choices that normalize and reward acceptance.

Differences in AI adoption transgressions across prompt categories further suggest that task framing plays a central role in shaping human–AI collaboration. Explanatory prompts, which emphasize correctness, justification, or general knowledge, exhibited the highest rates of attempted suggestion use. In contrast, creative and observational prompts, which are grounded in imagination or immediate perception, showed minimal AI adoption. One interpretation is that when users perceive a task as having an externally correct or authoritative answer, AI suggestions align with their expectations of assistance. When tasks are experiential or expressive, AI-generated continuations may offer less perceived value or even feel intrusive.

The game-based framing of \textit{Nonslop} was not intended to motivate performance or maximize engagement in the conventional sense, but to create a space where micro-decisions around AI use could be observed and even emphasized. By attaching visible consequences to suggestion use, the game transformed what is typically an invisible interaction into an explicit choice. This supports the use of lightweight game mechanics as a research lens for studying human–AI interaction, particularly in domains where default interface designs obscure moments of negotiation and agency.

The present study focused on short, self-contained writing tasks. Extending this framework to longer or cumulative writing activities, such as multi-paragraph narratives or sequential prompts, could reveal how AI adoption strategies evolve over time. With repeated exposure, users may renegotiate the balance between convenience and control as stylistic consistency and long-term voice become more relevant. A larger or more persistent participant pool would also enable analysis of within-user change, allowing future studies to distinguish novelty effects from learned strategies.

These findings have implications for the design of AI-assisted writing systems. Interfaces that default to frictionless suggestion acceptance may mask user ambivalence or preference for autonomy, particularly in creative or expressive tasks. Introducing optional friction, transparency, or contextual controls could allow users to more deliberately negotiate when and how AI assistance enters their writing process. While \textit{Nonslop} exaggerates these dynamics through a game framing, similar principles may apply to non-game interfaces that aim to support creativity without eroding individual voice.

This study is limited by its small sample size, short response lengths, and the technical constraints imposed by in-browser inference. The population able to access the game was skewed toward users with compatible hardware and browsers, and the playful framing may not generalize to professional writing contexts. Because the system was framed as a game, some behaviors may also reflect strategies for completing the task or avoiding penalties rather than stable preferences about creative voice. As such, the results should be interpreted as indicative rather than definitive. The strength of the approach lies not in representativeness but in its ability to surface behaviors that are difficult to observe in conventional AI-assisted writing environments.

\section{Conclusion}

This paper introduced \textit{Nonslop}, a small web-based game designed to study how people respond to AI-generated word suggestions when those suggestions are available but discouraged. By combining in-browser next-word prediction with explicit feedback on suggestion use, the game creates a setting where players must decide whether adopting AI-generated language is worth the potential penalty.

Across 214 submissions from 74 participants, we found that most users rarely attempted to adopt suggested words. In nearly three quarters of submissions, no AI suggestion transgressions were made at all. When we aggregated behavior at the user level, a clear pattern emerged: the majority of players were ``minimalists'' who produced short responses with almost no AI adoption transgressions, while a smaller group of ``selective adopters'' wrote longer responses and used suggestions occasionally, and an even smaller group of ``active adopters'' interacted more frequently and used suggestions more often. This suggests that, even under the same interface and incentive structure, users adopt distinct strategies for how to engage with AI support.

We also observed systematic differences across prompts. Explanatory prompts showed much higher average AI adoption transgressions than creative or observational prompts, and the most frequently answered prompts tended to have slightly lower suggestion use. Together, these results indicate that both task type and prompt appeal shape how likely users are to draw on AI-generated continuations. 

These findings are limited by the size and composition of our sample, as well as by the technical constraints of the Web-LLM setup, which excluded many mobile and locked-down systems. The responses themselves were short, and the prompts covered a narrow range of task types. Nonetheless, the study demonstrates that a simple game can make otherwise hidden decisions around AI suggestion transgressions observable and analyzable.

Future work could extend this approach by testing different incentive structures, comparing interfaces that block, warn about, or encourage AI use, and examining longer or more open-ended writing tasks. More broadly, this work supports the use of game-like environments as testbeds for understanding how people actually use, avoid, and negotiate with AI writing tools in practice. Future work should also compare the current local-inference implementation with a more technically robust version, such as a server-side or API-based suggestion system, to separate behavioral effects of the game design from access failures, latency, and browser compatibility issues.

\bibliographystyle{IEEEtran}

\bibliography{references}

\end{document}